\documentclass{zapiski}
\originfo{1}{}{}{2026}
\date{}

\usepackage[utf8]{inputenc}
\usepackage[T1]{fontenc}
\usepackage[english]{babel}

\usepackage{times}
\usepackage{url}
\usepackage{latexsym}
\usepackage{multirow}
\usepackage{graphicx}
\usepackage{amssymb}
\usepackage{hyperref}
\usepackage{booktabs}
\usepackage{tabularx}
\usepackage{caption}
\usepackage{cite}
\usepackage{paralist}
\usepackage{subcaption}
\usepackage{xcolor,colortbl}
\usepackage{array}
\usepackage{setspace}
\usepackage{placeins}
\usepackage{textcomp}
\usepackage{float}
\restylefloat{table}

\makeatletter
\AtBeginDocument{\let\@serieslogo\@empty}
\makeatother

\english

\begin{document}

\title[Recognizing Insufficient Audio Evidence]{Can Speech LLMs Recognize When Audio Evidence Is Insufficient? A Diagnostic Study and a Pre-Generation Probing Mitigation}

\author[Recognizing Insufficient Audio Evidence]{Vladislav Nesterenko\textsuperscript{1}, Polina Doronicheva\textsuperscript{2}, Olga Golubeva\textsuperscript{3}, Polina Sheviakova\textsuperscript{2}}
\address{\textsuperscript{1}Moscow Institute of Physics and Technology; \textsuperscript{2}HSE University; \textsuperscript{3}Chelyabinsk State University}
\email{nesterenko.vv@phystech.edu}

\begin{abstract}
Speech large language models are usually evaluated on questions that have answers. We ask what they do when a recording establishes none, and whether the standard way of finding out is sound. Diagnosed on synthesized speech the answer looks clear: Qwen2-Audio fabricates an answer to 92.5\% of unanswerable questions, while an ASR cascade reading the same recordings reaches 2.5\% hallucination at 6.7\% over-refusal, placing the failure in reasoning rather than in perception. The identical protocol on 376 native SQuAD~2.0 questions over human-read recordings removes that conclusion: the cascade hallucinates on 67.7\% against the end-to-end model's 69.8\%, at a higher over-refusal cost. A paired subset asking the same questions of a synthesized and a human recording of each passage localises the effect in the ASR stage --- the end-to-end model's abstention behaviour does not move with the speaker, the cascade's goes from 11.1\% to 65.9\%. Reliance on synthesized speech has been named as a limitation of audio benchmarks but not quantified; we quantify it, and show it large enough to reverse a finding. Since no prompt we tested repairs the residual failure, we transfer pre-generation attention probing to the multimodal setting. A probe over frozen prompt representations reaches 0.773 AUROC at predicting which unanswerable questions the model will answer anyway, where semantic entropy over ten sampled generations is at chance: the model fabricates the same answer every time, so sample disagreement cannot see a failure it is confident about.
\end{abstract}


\maketitle

\section{Introduction}
\label{sec:introduction}

Large Audio-Language Models, or Speech LLMs \cite{chu2024qwen2audio, radford2022whisper}, attach audio encoders directly to instruction-tuned text backbones, removing the latency of an ASR pipeline and performing well on spoken question answering \cite{sakshi2024mmau, wang2025mmsu}. Beneath that fluency lies a closed-world assumption: that every query is grounded in the audio. In deployment it often is not, and these models rarely abstain --- they produce fluent, unfounded answers \cite{zhao2026halluaudio, kuan2026aquabench} --- while telling a model it may say ``I don't know'' triggers indiscriminate over-refusal on questions the audio does answer \cite{ma2025reliable}.

This raises a diagnostic question: \textbf{do Speech LLMs hallucinate because they fail to perceive the signal, or because they fail to reason about what it establishes?} The natural instrument is an ASR cascade, which makes the same evidence textual: whatever a text model establishes from a clean transcript is information that survived the audio.

We first built a diagnostic set the way this field builds them --- synthesized speech, generated then human-verified questions --- and got a clean answer. Qwen2-Audio hallucinated on 92.5\% of unanswerable questions --- naming, for instance, Johann Gutenberg as the printer of Luther's 1521 writings --- and the abstention instruction cut this to 17.5\% while refusing 61.7\% of answerable ones, the model not distinguishing heard from unheard but simply abstaining more. The cascade reached 2.5\% hallucination at 6.7\% over-refusal. Since its transcript derives from the audio, the information needed to abstain is demonstrably in the signal, and the bottleneck would then be reasoning, not perception.

\textbf{The conclusion does not survive natural speech.} That set has two suspect properties: synthesized audio and model-written questions. The first is a known concern --- recent benchmarks name reliance on synthetic speech as a limitation and respond by building on human recordings \cite{lin2026vibe, hu2025vcbbench} --- but it has been avoided rather than measured. We measure it. On 376 native SQuAD~2.0 questions over recordings read by 40 speakers, the cascade's advantage vanishes: 67.7\% hallucination against the end-to-end model's 69.8\%, intervals overlapping, at a higher over-refusal cost. A paired subset in which the same 214 questions are asked over both a synthesized and a human recording of each passage localises the effect: the end-to-end model's abstention behaviour is statistically unchanged by the audio source, while the cascade moves from 11.1\% hallucination to 65.9\% ($p < 10^{-12}$). Synthesized audio is easier for everything that listens to it, but only a transcription stage converts that ease into apparent epistemic competence.

Both pipelines therefore fail on natural speech, and the abstention instruction does not repair it. We therefore transfer pre-generation attention probing \cite{miftakhova2026probing} to the multimodal setting, reading frozen prompt representations rather than sampling the output \cite{kuan2026uncertainty}.

Our contributions are three-fold. \textbf{(i) A paired protocol isolating the audio source:} a twin subset asking identical questions over the same passages in both conditions; we are aware of no published resource containing both halves of such a pair. \textbf{(ii) A measurement of how far diagnostic conclusions transfer:} reliance on synthesized speech has been named as a limitation but not quantified; we quantify it, exhibit a case where it reverses a finding, and separate it from question provenance, which drives the other system entirely. \textbf{(iii) Pre-generation detection of unanswerability:} a probe over frozen hidden states identifies, before any token is generated, which unanswerable questions the model will answer anyway --- the signal an abstention gate needs --- where sampling-based uncertainty over the same generations is at chance.

\section{Related Work}
\label{sec:related_work}

\textbf{Unanswerability in audio QA.} Standard protocols \cite{sakshi2024mmau, huang2024dynamicsuperb} assume every question is grounded in the audio. Two recent benchmarks drop that assumption: AQUA-Bench \cite{kuan2026aquabench} and HalluAudio \cite{zhao2026halluaudio}. Neither can host the experiment below. AQUA-Bench contains no speech at all, so a cascade is undefined on it; HalluAudio does cover speech, but its speech items probe whether the model \emph{heard} correctly rather than whether it can tell that a passage establishes no answer. Neither includes a cascade control, and neither varies the audio source. Appendix~\ref{app:benchmarks} gives the composition of both. We work instead with questions requiring semantic inference over multi-fact spoken passages, unanswerability subtyped by cause.

\textbf{Synthesized speech as a confound.} The field has flagged this. VIBE \cite{lin2026vibe} observes that existing speech fairness benchmarks ``rely on synthetic speech and Multiple-Choice Questions'' and evaluates on human recordings instead; VCB Bench \cite{hu2025vcbbench} likewise names synthetic speech as a limitation and is built on real human speech. Both \emph{avoid} the confound; neither measures it. We quantify it in a paired design holding the questions fixed, and show it large enough to reverse a conclusion.

\textbf{Cascades and mitigation.} Cascaded pipelines are the standard comparison point in spoken QA, but the comparison is about answer accuracy or latency \cite{lin2022dual}; Jung and Choi \cite{jung2026korean} show cascade degradation tracks information lost at the ASR stage, the mechanism our results expose from the other side. We are not aware of prior work comparing the two architectures on abstention. Mitigations either train the model --- chain-of-thought prompting and fine-tuning on ``I don't know'' data \cite{ma2025reliable}, contrastive alignment and synthesized negatives \cite{kuan2025balsa, kuan2025listen} --- or elicit uncertainty \emph{after} generation, at a cost of $k$ generations before the system may speak \cite{kuan2026uncertainty}. Closest to ours is AEG \cite{yang2026aeg}, which also reads a Speech LLM's cross-modal attention and compares against a cascade, but fine-tunes the model to calibrate that attention and targets answer quality on answerable questions. We fit a small external probe on frozen pre-generation representations and use it to decide whether to answer at all.

\section{Task and Data}
\label{sec:data}

A system receives one recording and one written question and must either answer or abstain. We offer no answer options and ask for no confidence score: the decision to speak is the object of study, so it is taken in the same free-form channel as the answer.

Each item carries a category fixing the correct behaviour. In \textbf{A (stated)} the answer is spoken in the recording; in \textbf{B (inferable)} it follows from what is spoken without being said verbatim; in \textbf{C (unanswerable)} the recording establishes no answer and the system should abstain. C is subtyped by \emph{why} the question fails: the recording never mentions the entity asked about, mentions it but not the attribute asked for, presupposes something it never establishes, or is unrelated. The middle two matter most --- there the recording supplies enough surface overlap for a plausible answer, and a model keying on topical similarity rather than evidence will compose one. Answering a C item is a \textbf{hallucination}, abstaining on an A or B item an \textbf{over-refusal}; any mitigation trades one against the other.

\textbf{Diagnostic set.} Built on Spoken-SQuAD \cite{li2018spokensquad}, whose audio is generated from SQuAD paragraphs by a text-to-speech system. From 40 hand-selected passages we keep one native SQuAD question each for A and generate B and C items from the transcripts, filtered and then verified by annotators who had not written them. Frozen at 100 items: 30 A, 30 B, 40 C (Appendix~\ref{app:pilot}).

\textbf{Validation set.} It removes both suspect properties at once: NMSQA \cite{lin2022dual}, in which SQuAD paragraphs are read aloud by human speakers, intersected with SQuAD~2.0 \cite{rajpurkar2018squad2}, whose unanswerable questions were written by crowdworkers instructed to make them \emph{look} answerable against the paragraph before them. Every recording is trimmed to 30 seconds, since Qwen2-Audio processes only that much while the cascade receives the whole transcript, and an A question is kept only if its answer falls inside the audible span (Appendix~\ref{app:scale}). The frozen set is \textbf{376 items, 141 A and 235 C}, over 48 paragraphs, 51 recordings and 40 speakers. It has no category B: SQuAD has no native inferable questions, and generating them would reintroduce the property this set exists to eliminate.

\textbf{Twin subset.} Of those paragraphs, 26 are ones for which we also hold synthesized audio. They carry the identical 214 questions (88 A, 126 C) in both conditions, with audible content matched to within a word of median transcript length (Appendix~\ref{app:scale}), so the subset varies the audio source and nothing else.

\section{Methods}
\label{sec:methods}

\subsection{Systems and prompts}

Two systems, identical audio, both in 8-bit quantisation with greedy decoding. The \textbf{end-to-end Speech LLM} is Qwen2-Audio-7B-Instruct \cite{chu2024qwen2audio}; the \textbf{cascade} is Whisper-medium \cite{radford2022whisper} transcribing the recording, followed by Qwen2.5-7B-Instruct \cite{qwen2024qwen25} answering from the transcript alone. The cascade is an instrument, not a competitor: it sees the same evidence in explicitly textual form, so if it abstains correctly where the end-to-end model does not, the failure lies downstream of perception.

Two prompts, fixed throughout. The \textbf{plain} prompt says nothing about abstention; the \textbf{IDK} prompt adds one sentence licensing it and fixing its surface form, which also keeps the abstention detector honest. This is the cheapest mitigation available --- one sentence, no training, no extra decoding --- which is why it is the baseline worth beating (Appendix~\ref{app:prompts}).

\subsection{Grading and metrics}

A rule-based classifier labels each response \texttt{answer}, \texttt{abstain} or \texttt{hedge}; substantive answers on answerable categories are then graded by a language-model judge, since rules alone false-negative paraphrase such as ``BSkyB'' against a gold ``British Sky Broadcasting''. Unanswerable items are excluded from judging: there only the label matters, any substantive answer being a hallucination whatever it says. The judge is Qwen3-8B without step-by-step reasoning, chosen from seven candidates against 93 hand-graded answers and agreeing with them on 95\% (Appendix~\ref{app:judge}).

On unanswerable items we report the \textbf{hallucination rate} (share answered substantively) and the correct-abstention rate, hedges counted separately. On answerable items we report \textbf{accuracy} over \emph{all} items of the category, so a refusal counts against accuracy exactly as a wrong answer does, and the \textbf{over-refusal rate}. Proportions carry 95\% Wilson intervals; the twin subset is a paired design, so differences there are tested with McNemar's test on the discordant items.

\subsection{Pre-generation probing}
\label{sec:methods_probing}

Following \cite{miftakhova2026probing}, we probe representations taken \emph{before} any token is generated, so detection costs one forward pass rather than $k$ sampled generations. Hidden states are captured from the full multimodal prompt at four candidate layers under four poolings, and a ridge probe predicts the model's own error frequency --- the fraction of $k=10$ completions, sampled at the model's own temperature, that are wrong. Regression rather than classification, since the target is a fraction; cross-validation grouped by passage, since items sharing a passage share audio. The probe is external and the weights are untouched, so its intended use is as an abstention gate consulted before decoding; what we measure here is the detection quality such a gate would rest on, not the gate's operating curve (Appendix~\ref{app:probing}).

We report AUROC overall \emph{and} within the unanswerable category. The restriction is the honest measurement: answerable and unanswerable items differ systematically in difficulty, so a probe that had learned only which category an item belongs to would score well overall and be useless for the decision that matters.

\section{Experiments and Results}
\label{sec:experiments}

Four configurations --- two systems $\times$ two prompts --- on each set, each a single greedy pass, so intervals are binomial. Full per-run tables are in Appendix~\ref{app:tables}; the diagnostic set's headline figures appear in Section~\ref{sec:introduction}.

\subsection{Natural speech: the dissociation disappears}

On the validation set the cascade's advantage is gone. Under the abstention instruction it hallucinates on 67.7\% of unanswerable questions against the end-to-end model's 69.8\% --- overlapping Wilson intervals, where the diagnostic set showed 2.5\% against 17.5\% --- and pays more for that parity, 11.3\% over-refusal against 0.7\%; under plain prompting the two sit at 87.2\% and 94.0\%. The instruction has lost most of its force, removing a quarter of the end-to-end model's hallucinations where it removed three quarters. Under the best prompt available to them, both systems still answer more than two thirds of questions that have none.

The collapse is traceable to the front-end: the cascade's accuracy on stated facts falls from 100\% to 72.3\%, errors concentrated where ASR is weakest --- ``Monterey'' transcribed as ``Ontario'', ``Khanbaliq'' as ``Can Balik'' --- matching Jung and Choi \cite{jung2026korean}, who find cascade degradation tracking information lost at the ASR stage.

\subsection{Isolating the audio source}
\label{sec:twin}

The two sets differ in audio source \emph{and} question provenance, so neither alone attributes the change to synthesis. The twin subset varies only the first, and because the design is paired we test it with McNemar over the discordant items.

Synthesized speech is easier for both systems to answer from --- accuracy is higher under synthesis in all four configurations. What separates them is whether that advantage extends to knowing when \emph{not} to answer. The end-to-end model's abstention behaviour does not move with the speaker: 92.1\% hallucination against 92.9\% under plain prompting ($p = 1.00$), 78.6\% against 72.2\% under the abstention instruction ($p = 0.10$, and in the direction opposite to what an artifact account predicts). The cascade's is transformed: the same questions take it from 11.1\% to 65.9\% ($p < 10^{-12}$, 72 discordant items one way against 3 the other) and from 64.3\% to 87.3\% under plain prompting ($p = 4\cdot10^{-6}$), while its accuracy falls by 22.7 points against the end-to-end model's 9.1.

Both face identical questions, passages and audible content --- and both recordings are trimmed to 30 seconds, so the cascade's lead here is not the artifact of it receiving a transcript of the whole file, as it partly could have been on the untrimmed diagnostic set. The difference is attributable to the one component the two systems do not share. The dissociation seen on the diagnostic set was therefore not evidence about perception versus reasoning, but a measurement of how much easier synthesized speech is to transcribe.

Because the twin subset shares its audio with the diagnostic set and its questions with the validation set, placing the three side by side separates the two confounds: for the end-to-end model the entire effect lies in question provenance (61.1 points) and none in audio, while for the cascade it is the reverse (54.8 points from audio). The two cautions of this paper are thus independent findings with distinct mechanisms, each demonstrated by the system it applies to (Appendix~\ref{app:decomp}).

\subsection{Pre-generation probing}
\label{sec:probing_results}

Probes are fitted and evaluated on the unanswerable items alone --- stricter than mixing the categories, since with only one present a probe cannot score well by recognising which category an item belongs to. The number it earns is then the one that matters: given a question the audio does not answer, will this model answer it anyway?

It can be predicted. The last prompt position at layer 16 gives \textbf{0.773} AUROC (95\% bootstrap 0.671--0.863). The candidate layers were fixed before any result was seen, on the finding that middle layers carry most of this signal in text-only models \cite{miftakhova2026probing}; layer 16 is where that prediction lands, and layers 8 and 32 bracket it as expected. Pooling over audio positions carries nothing usable at any layer: the risk of fabricating an answer is legible in how the model has integrated the question, not in its encoding of the recording (Appendix~\ref{app:probing}).

\textbf{The post-generation baseline fails on this task, and instructively.} Semantic entropy over the same ten samples scores 0.506 (0.405--0.600), chance, with an interval that does not reach the probe's. The reason is visible in the labels: the model returns the same verdict on all ten samples for 92\% of unanswerable items, fabricating the same answer every time. Sampling disagreement can only detect a failure the model is unsure about, and this one it is sure about. The internal state distinguishes these items when the output distribution does not --- which is the case for looking before generation rather than after it, and not merely the latency argument we began with.

\section{Limitations}
\label{sec:limitations}

Four things bound what these numbers support. The diagnostic set has 100 items, and the probing evaluation rests on 25 negative items out of 235, so several intervals here are wide and the single best probe configuration --- one of sixteen evaluated --- carries selection optimism. Both systems are built from Qwen models and the synthesized side of the twin pair comes from one text-to-speech system, so magnitudes are specific to this setup even where directions should not be. Correctness on answerable items is decided by a model agreeing with human graders on 95\% of a calibration sample, which is a floor on every accuracy figure. And only two prompts were tested, so ``no prompt repairs this'' is a statement about the abstention instruction rather than about prompting in general. Appendix~\ref{app:limits} gives these in full, along with the 30-second encoder window and the scope of the data.

\section{Discussion and Conclusion}
\label{sec:discussion}

\textbf{A diagnostic set can be wrong about the thing it was built to diagnose.} Ours was built carefully --- hand-selected passages, a structured taxonomy, verification by annotators who had not written the items, an explicit freeze --- and still supported a conclusion the validation set overturned. The flaw lay not in the verification but in two properties of the material that verification does not test: neither is visible when checking whether an individual item is correctly labelled, and both determine what the aggregate can be used to conclude.

\textbf{The failure is systemic, and prompting is the wrong lever.} On natural speech both pipelines fail at rates whose intervals overlap, so a clean textual view of the same evidence does not repair the failure --- which removes a whole class of proposed fixes, a better front-end in front of a text model. The abstention instruction is one fixed operating point: it makes a system likelier to abstain everywhere, indiscriminately, and cannot condition on the item, because the model it instructs cannot tell the two kinds of item apart --- the original failure restated. Post-generation uncertainty estimation does condition on the item, at a cost of $k$ generations before the system may speak --- in a voice interface, undoing the latency advantage for which an end-to-end model was chosen. Probing the prompt representation asks whether the same decision can be made from state already computed, and here it can: 0.773 AUROC against chance for the sampling baseline. The reason that baseline fails is the more interesting half of the result. It assumes an uncertain model betrays itself by answering differently each time, and this model does not --- on 92\% of unanswerable items it fabricates the same answer on all ten samples. Confident fabrication is invisible to output-level uncertainty and legible in the representation that produced it. The case for looking before generation is therefore not only that it is cheaper. Data, manifests, prompts, grading pipeline and run reports are released at \url{https://github.com/ladnlav/slm-audio-evidence}, so that both the original finding and its refutation can be rechecked.

\section*{Contribution}
\label{sec:contribution}

\textbf{Vladislav Nesterenko} posed the research question and designed the study: the two-system comparison, the taxonomy of unanswerability, and the protocol by which the diagnostic finding was tested against natural speech. Implemented the pre-generation probing method, analysed all results, and wrote the paper.

\textbf{Polina Doronicheva} built and validated both datasets, from source selection and question construction through verification to the frozen releases, and established which existing benchmarks could and could not serve as validation. Ran the inference experiments.

\textbf{Olga Golubeva} surveyed the literature on hallucination benchmarks and abstention mitigation for audio-language models, positioned the work against it, and drafted the Introduction and Related Work.

\textbf{Polina Sheviakova} designed, calibrated and validated the automatic grading pipeline against human judgement --- the instrument behind every correctness figure reported here --- and drafted its description.

\section*{Acknowledgments}
This work was carried out at SMILES~2026, the Summer School of Machine Learning at the Skolkovo Institute of Science and Technology. We thank Assel Yermekova for mentoring the project during the summer school.

\bibliographystyle{amsplain}
\bibliography{references}

\appendix

\section{Why the existing unanswerability benchmarks do not host this experiment}
\label{app:benchmarks}

\textbf{AQUA-Bench} \cite{kuan2026aquabench} evaluates three unanswerability scenarios --- a missing correct option, a categorically mismatched option set, and a question not grounded in the audio. Its audio is drawn from ESC-50 animal sounds, a musical-instrument set, non-verbal human vocalisations and MMAU: no speech, and therefore no transcript for a cascade to reason over. Every item is multiple-choice, with abstention expressed by selecting ``none of the above''. That format also bounds the failure being measured: a model must pick an option, so it cannot exhibit the free-form fabrication a voice assistant produces, and being offered a ``none'' option is itself a hint.

\textbf{HalluAudio} \cite{zhao2026halluaudio} is broader --- over 5{,}000 human-verified pairs across speech, environmental sound and music, with open-ended items alongside multiple-choice ones, and a reported refusal rate. Its speech subset is built on Common Voice, and its speech tasks ask whether the model heard correctly: word-level alignment under deletion or substitution, counts of a spoken word, and paralinguistic attributes such as speaker gender or speaking rate. These are perception probes over short single-sentence clips rather than semantic questions over multi-fact passages, so an item can be answered without any judgement about what the passage establishes. Neither benchmark reports a cascade comparison, and neither varies whether its audio was spoken by a machine or a person.

\section{Prompts}
\label{app:prompts}

Answering prompts, used verbatim for both systems. \texttt{\{question\}} is substituted per item; the cascade receives the same text preceded by \texttt{The audio is given to you as a transcript. Transcript: "..."}.

\begin{quote}\small\ttfamily
\textbf{plain:} Listen to the audio and answer the question. Question: \{question\}
\end{quote}

\begin{quote}\small\ttfamily
\textbf{IDK:} Listen to the audio and answer the question using ONLY information from the audio. If the audio does not contain the information needed, reply exactly: "The audio does not provide that information." Question: \{question\}
\end{quote}

Judge prompt, applied to substantive answers in categories A and B:

\begin{quote}\small\ttfamily
You grade a QA system. TRANSCRIPT: \{transcript\} QUESTION: \{question\} GOLD ANSWER: \{gold\} SYSTEM ANSWER: \{response\}. Reply with exactly one word: CORRECT (if the system answer has the same meaning as the gold answer), INCORRECT (different, wrong, or contradicting), or ABSTAINED (if the system declines to answer or says there is no information). Judge meaning, not the exact wording.
\end{quote}

\section{Diagnostic set construction}
\label{app:pilot}

Source audio is Spoken-SQuAD \cite{li2018spokensquad}, in which SQuAD paragraphs are rendered to speech by a text-to-speech system. We selected 40 passages by hand, each required to be 20--60 seconds long (in the event 31--55 s, median 43 s) and to contain at least three concrete facts. Category A reuses one native SQuAD question per passage. Categories B and C were generated from the passage transcript with one templated prompt per subtype.

Generated candidates passed an automatic filter that rejected yes/no questions, length violations, within-passage duplicates, B items whose answer proved to be stated verbatim, and C items that re-used passage keywords without introducing a new focus. Of 440 candidates (120 B, 320 C), 423 survived.

Verification was performed by two annotators who had not written the items --- 63 and 62 rows respectively, with 20 shared calibration items --- each checking category and subtype assignment, inference validity for B, unanswerability from the transcript for C, and naturalness. Agreement on the shared items was 80\% (16/20); the four disputed items were dropped by rule rather than adjudicated, and three C items were kept after rewording. The set was then frozen at 100 items: 30 A, 30 B, and 40 C (12 absent-entity, 12 missing-attribute, 8 false-presupposition, 8 off-topic). It has not been modified since, so all four runs are comparable.

\section{Validation set construction}
\label{app:scale}

NMSQA \cite{lin2022dual} distributes each paragraph as a sequence of numbered segments rather than as one recording. A single segment cannot stand in for the paragraph, since a question may depend on material in another, so we concatenate the segments of each paragraph in order, leaving the source files untouched.

Every recording is then trimmed to 30 seconds, with the transcript truncated proportionally, and an A question is kept only if its answer falls inside the audible span. This equalises the evidence available to the two systems: Qwen2-Audio processes only the first 30 seconds of a recording, while the cascade receives a transcript of the whole file. C questions are unaffected --- removing material from the end of a recording cannot make an unanswerable question answerable. The result is 145 A and 235 C items over 48 paragraphs, 51 recordings and 40 speakers.

A final pass listened to the A items whose gold answer an automatic transcript check had flagged. Four were dropped: three because the answer is not in fact audible within the 30-second span, and one because a fuzzy match was likely spurious (``Social Charter'' against a gold ``Social Chapter''). Eight were listened to and kept, because there the answer is clearly audible and the flag came from the ASR front-end mangling a proper name --- ``Monterey'' transcribed as ``Ontario'', ``Khanbaliq'' as ``Can Balik'', ``Jean Ribault'' as ``Rebolt'', ``Aare'' as ``Arriff''. Dropping those would have deleted precisely the cases on which the cascade is expected to fail, biasing the comparison in its favour. The frozen set is 376 items: 141 A and 235 C.

\section{Judge selection and calibration}
\label{app:judge}

Seven candidate models and two prompt formulations were compared against 93 category-B answers graded by hand (Table~\ref{tab:judge}). We selected Qwen3-8B without step-by-step reasoning: its agreement is within one item of the best candidate, it is open-weight, and at 1.09~s per response it is roughly three times faster than the next-best-agreeing model of comparable quality.

\begin{table}[htbp]
\centering\small
\caption{Judge selection: agreement with human grading on 93 category-B answers.}
\label{tab:judge}
\begin{tabular}{lcc}
\toprule
Model & Agreement & Time/response \\
\midrule
Gemma-4-12B & 96\% (89/93) & 1.70 s \\
\textbf{Qwen3-8B} & \textbf{95\% (88/93)} & \textbf{1.09 s} \\
Qwen3.6-27B & 95\% (88/93) & 3.09 s \\
Ministral-3-8B-Instruct & 92\% (86/93) & 1.99 s \\
Qwen3.5-9B & 92\% (86/93) & 1.89 s \\
Qwen3-8B, step-by-step rubric & 88\% (82/93) & 15.78 s \\
Ministral-3-8B, step-by-step rubric & 84\% (78/93) & 52.44 s \\
\bottomrule
\end{tabular}
\end{table}

Three plausible improvements were tested and all three rejected on the numbers. Instructing the judge not to reward verbose answers \emph{lowered} agreement from 92\% to 89\%: the model began treating length as evidence of correctness instead of checking the fact. Removing the transcript from the judge's input --- on the reasoning that the rubric never refers to it --- lowered agreement from 94\% to 90\%; the transcript stabilises the decision even though nothing in the rubric mentions it. Enabling step-by-step reasoning did not improve agreement, cost an order of magnitude in time, and left nearly a fifth of responses with no parseable verdict.

Since judge--human disagreement need not be the judge's fault, we inspected the disagreements by hand and found two defective gold answers. In one item the gold listed only one of the two properties named in the recording, so every answer naming both was marked wrong; the gold was completed. In another, an answer matching all five gold categories verbatim was marked wrong while a near-identical answer in a different run had been accepted; the label was corrected. After both fixes, agreement rose from 87/93 to 88/93.

\paragraph{A scope defect worth recording.} Our first implementation invoked the judge for category B only --- a rule written when B was the only category whose grading was thought to need semantics. Because the validation set contains no B items, the judge never fired there, and all 376 items were graded by string comparison alone. The effect was not cosmetic: paraphrases were counted as errors at scale (``Oxygen therapy'' against a gold ``oxygen supplementation''; ``thirty-two degrees Celsius'' against ``32~\textdegree C''), and it hit the cascade hardest, since re-transcribed answers are reworded more often. Correcting the scope and re-grading raised category-A accuracy by up to 30 percentage points on the validation set and by 13 points on the pilot. All numbers in this paper are post-correction.

\section{The 30-second window}
\label{app:window}

Qwen2-Audio's audio encoder has 1500 positional embeddings, covering 3000 mel frames or 30 seconds; the limit is architectural rather than a preprocessing choice, and we confirmed it by running the processor on 43-second and 90-second inputs and observing exactly 30 seconds of signal in both cases. The cascade, receiving a Whisper transcript of the whole file, has no equivalent bound, so on longer recordings the two systems do not see the same evidence. The validation and twin sets are therefore trimmed to 30 seconds with answerable questions filtered for audibility.

The diagnostic set predates this rule and is untrimmed, its passages running 31--55 seconds (median 43). This is what depresses the end-to-end model's accuracy on stated facts there --- 33.3\%, against 80.1\% on the trimmed validation set and 90.9\% on trimmed synthesized audio in the twin subset. Since the twin comparison rules out synthesis as the cause, the untrimmed length is what remains. Of the 24 diagnostic-set A items whose gold answer appears verbatim in the transcript, one falls beyond the audible span; no equivalent check is possible for category B, whose answers are by definition not verbatim, and that residual risk is why the diagnostic set carries no argument the validation set does not also carry.

\section{Full result tables}
\label{app:tables}

\begin{table}[htbp]
\centering\small
\caption{Diagnostic set (synthesized audio): 30 A, 30 B, 40 C. 95\% Wilson intervals. Correct abstention on unanswerable items is 2.5, 82.5, 27.5 and 97.5\%, the remainder being hedges: 5.0, 0.0, 10.0 and 0.0\%.}
\label{tab:pilot}
\setlength{\tabcolsep}{4pt}
\begin{tabular}{llcccc}
\toprule
System & Prompt & Halluc.\ $\downarrow$ & Acc.\ A $\uparrow$ & Acc.\ B $\uparrow$ & Over-ref.\ $\downarrow$ \\
\midrule
Qwen2-Audio & plain & 92.5 \tiny(80.1--97.4) & 33.3 & 26.7 & 6.7 \\
Qwen2-Audio & IDK   & 17.5 \tiny(8.7--32.0) & 23.3 & 6.7 & \textbf{61.7} \\
Cascade     & plain & 62.5 \tiny(47.0--75.8) & 100.0 & 93.3 & 0.0 \\
Cascade     & IDK   & \textbf{2.5} \tiny(0.4--12.9) & 93.3 & 83.3 & 6.7 \\
\bottomrule
\end{tabular}
\end{table}

Category B is graded by hand against the gold answer and transcript; it exists only in this set.

\begin{table}[htbp]
\centering\small
\caption{Validation set (natural speech, native SQuAD~2.0): 141 A, 235 C. Hallucination is over the unanswerable items, accuracy and over-refusal over the answerable ones; 95\% Wilson intervals. Hedged responses (0.4, 0.0, 5.1 and 0.0\%) count as neither answer nor abstention. No category B by construction.}
\label{tab:scale}
\setlength{\tabcolsep}{4pt}
\begin{tabular}{llccc}
\toprule
System & Prompt & Hallucination $\downarrow$ & Accuracy $\uparrow$ & Over-refusal $\downarrow$ \\
\midrule
Qwen2-Audio & plain & 94.0 \tiny(90.2--96.4) & \textbf{80.1} \tiny(72.8--85.9) & 0.0 \\
Qwen2-Audio & IDK   & 69.8 \tiny(63.6--75.3) & 79.4 \tiny(72.0--85.3) & 0.7 \\
Cascade     & plain & 87.2 \tiny(82.4--90.9) & 72.3 \tiny(64.4--79.1) & 0.7 \\
Cascade     & IDK   & 67.7 \tiny(61.4--73.3) & 68.8 \tiny(60.7--75.9) & \textbf{11.3} \\
\bottomrule
\end{tabular}
\end{table}

\begin{table}[htbp]
\centering\small
\caption{Twin subset: identical questions in both audio conditions, both sides graded by the same judge. Hallucination over the 126 unanswerable items, accuracy over the 88 answerable ones; $p$ is two-sided McNemar over the discordant items. Rebuilt by \texttt{scripts/twin\_report.py}.}
\label{tab:twin}
\setlength{\tabcolsep}{3.5pt}
\begin{tabular}{llccc|ccc}
\toprule
& & \multicolumn{3}{c|}{Hallucination $\downarrow$} & \multicolumn{3}{c}{Accuracy $\uparrow$} \\
System & Prompt & synth & human & $p$ & synth & human & $p$ \\
\midrule
Qwen2-Audio & plain & 92.1 & 92.9 & 1.00 & 90.9 & 81.8 & 0.013 \\
Qwen2-Audio & IDK   & 78.6 & 72.2 & 0.10 & 87.5 & 79.5 & 0.023 \\
Cascade     & plain & 64.3 & 87.3 & $4\cdot10^{-6}$ & 94.3 & 71.6 & $5\cdot10^{-5}$ \\
Cascade     & IDK   & \textbf{11.1} & \textbf{65.9} & $<10^{-12}$ & 86.4 & 71.6 & 0.012 \\
\bottomrule
\end{tabular}
\end{table}

\section{Probing setup}
\label{app:probing}

Representations are captured from the full multimodal prompt --- audio plus question, exactly as the model receives it --- at layers 8, 16, 24 and 32. Audio and text occupy distinct spans of the sequence, so we record four views per layer: the hidden state at the last prompt position, the mean over all positions, the mean over audio positions, and the mean over text positions. Capturing every candidate layer and pooling in a single pass makes the choice of layer an analysis performed after the fact rather than a guess made in advance.

Soft targets come from $k=10$ completions sampled at temperature 1.0, the target being the fraction that are wrong: on category A a substantive answer graded incorrect, on category C any substantive answer. Sampling at the model's own temperature is required rather than incidental --- the unbiasedness argument for this estimator is stated for the model's own output distribution, so another temperature would estimate the error rate of a different distribution.

Probes are ridge regressions on standardised features, fitted with cross-validation grouped by source passage. The grouping is not a formality: items sharing a passage share audio, and on our data an ungrouped split reports AUROC near 1.0 where a grouped split reports approximately chance.

\begin{table}[htbp]
\centering\small
\caption{AUROC on the 235 unanswerable items, four candidate layers by four poolings, with 95\% bootstrap intervals for the best cell of each pooling. Middle layers lead at the last prompt position; audio-position pooling is at or below chance throughout. Semantic entropy over the same ten samples scores 0.506 (0.405, 0.600). Rebuilt by \texttt{scripts/train\_probe.py} and \texttt{scripts/entropy\_baseline.py}.}
\label{tab:probe}
\setlength{\tabcolsep}{4pt}
\begin{tabular}{lcccc@{\quad}l}
\toprule
Pooling & L8 & L16 & L24 & L32 & 95\% CI (best) \\
\midrule
Last prompt position & 0.580 & \textbf{0.773} & 0.733 & 0.771 & (0.671, 0.863) \\
Mean, text positions & 0.709 & 0.698 & 0.695 & 0.747 & (0.616, 0.860) \\
Mean, all positions & 0.466 & 0.555 & 0.698 & 0.707 & (0.598, 0.812) \\
Mean, audio positions & 0.319 & 0.568 & 0.539 & 0.484 & (0.465, 0.675) \\
\bottomrule
\end{tabular}
\end{table}

Two caveats on this grid. Sixteen configurations were evaluated and the best is reported, so that single figure carries selection optimism; the pattern is what should be trusted, and it is consistent --- the last prompt position at layers 16--32 and the text-position mean at layer 32 all land between 0.73 and 0.77. And the evaluation rests on 25 negative items out of 235, which is what makes the bootstrap intervals as wide as they are.

Category A was excluded from probe training because its soft labels require a judge verdict per sample --- 1{,}410 calls we did not run --- and grading those samples by string matching alone would have written paraphrases into the training signal as fabrications. Restricting to unanswerable items also removes the category-membership shortcut by construction, which is why we report no overall AUROC.

\section{Separating the two confounds}
\label{app:decomp}

The twin subset shares its audio with the diagnostic set and its questions with the validation set, so the three together move one factor at a time (Table~\ref{tab:decomp}).

\begin{table}[htbp]
\centering\small
\caption{Hallucination under the abstention instruction, one factor at a time. The middle column shares audio with the left and questions with the right. The second step is the controlled twin pair; the first also varies passages and trimming.}
\label{tab:decomp}
\setlength{\tabcolsep}{4pt}
\begin{tabular}{lccc|cc}
\toprule
& generated Q & native Q & native Q & \multicolumn{2}{c}{attributable to} \\
System & synth audio & synth audio & human audio & questions & audio \\
\midrule
Qwen2-Audio & 17.5 & 78.6 & 72.2 & $\mathbf{+61.1}$ & $-6.3$ \\
Cascade     & 2.5  & 11.1 & 65.9 & $+8.6$ & $\mathbf{+54.8}$ \\
\bottomrule
\end{tabular}
\end{table}

The two confounds separate, and each attaches to a different system. For the end-to-end model the whole effect lies in question provenance --- replacing generated questions with native adversarial ones costs 61.1 points, while audio contributes nothing. For the cascade it is the reverse, 54.8 points from audio against 8.6 from questions. Our two cautions are therefore not two symptoms of one muddle but independent findings with distinct mechanisms, each demonstrated by the system it applies to.

\section{Limitations in full}
\label{app:limits}

\textbf{Scale.} The diagnostic set has 100 items, so its intervals are wide --- the one on 92.5\% spans 80.1--97.4\% --- and per-subtype counts of 8--12 support description, not inference.

\textbf{One model family, one synthesizer.} Both systems are built from Qwen models of comparable size, and the synthesized side of the twin pair comes from a single text-to-speech system. The direction of the effect should hold for any synthesizer whose output is more predictable than human speech, but its magnitude is specific to this one. The 30-second encoder window is a further asymmetry between the two systems, discussed in Appendix~\ref{app:window}.

\textbf{Automatic grading.} Correctness on answerable items is decided by a model agreeing with human graders on 95\% of a 93-item sample --- a floor on every accuracy figure here --- drawn from a different category and audio source than where the judge is then applied.

\textbf{Scope of the claims.} All audio is English and read from written text rather than spontaneous. Only two prompts were tested, so ``no prompt repairs this'' is a statement about the abstention instruction, not about prompting in general. And the first step of Table~\ref{tab:decomp} varies passages and trimming alongside question provenance, so that step is indicative where the twin comparison is controlled.

\end{document}